\documentclass[11pt,a4paper]{article}
\usepackage[hyperref]{acl2019}
\usepackage{times}
\usepackage{latexsym}
\usepackage{tabularx}
\usepackage{graphicx}
\usepackage{multirow}
\usepackage{booktabs}
\usepackage{enumitem}
\usepackage{amsmath}
\usepackage{todonotes}
\usepackage{pifont}
\usepackage{amssymb}
\usepackage{color, colortbl}
\usepackage{url}
\usepackage{float}
\usepackage{appendix}
\usepackage{bbold}
\usepackage[T1]{fontenc}

\aclfinalcopy 
\def\aclpaperid{1967} 

\newcommand{\comet}{{\textbb{COMET}}}

\newcommand\cometemoji{\raisebox{-2pt}{\includegraphics[width=0.9em]{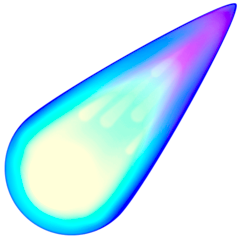}}}

\definecolor{Gray}{gray}{0.935}

\newcommand\modelname{\textbb{COMET}}

\title{
\comet{}\cometemoji{}: Commonsense Transformers \\for Automatic Knowledge Graph Construction
}

\newcommand\aitwo{$^\diamondsuit$}
\newcommand\uw{$^\spadesuit$}
\newcommand\msr{$^\clubsuit$}
\newcommand\aspace{\hspace{.75em}}

\author{
  Antoine Bosselut \aitwo\uw\aspace
  Hannah Rashkin \aitwo\uw\aspace
  Maarten Sap \aitwo\uw\aspace
  Chaitanya Malaviya \aitwo\aspace \\
  \textbf{Asli Celikyilmaz \msr\aspace
  Yejin Choi \aitwo\uw} \\
 \aitwo Allen Institute for Artificial Intelligence, Seattle, WA, USA\\
 \uw Paul G. Allen School of Computer Science \& Engineering, Seattle, WA, USA\\
 \msr Microsoft Research, Redmond, WA, USA
}

\date{}

\begin{document}
\maketitle
\begin{abstract}
We present the {first} comprehensive study on automatic knowledge base construction for two prevalent commonsense knowledge graphs: \textsc{Atomic} \cite{sap2018atomic} and {ConceptNet} \cite{speer2017conceptnet}.
Contrary to many conventional KBs that store knowledge with canonical templates, 
commonsense KBs 
only store loosely structured open-text descriptions of knowledge. We posit that an important step toward automatic commonsense completion is the development of \emph{generative} models of commonsense knowledge, and propose \emph{\textbf{COM}mons\textbf{E}nse \textbf{T}ransformers} (\modelname{}\cometemoji{}) that learn to {generate} rich and diverse commonsense descriptions in natural language. 
Despite the challenges of commonsense modeling, our investigation reveals promising results when implicit knowledge from deep pre-trained language models is transferred to generate explicit knowledge in commonsense knowledge graphs. Empirical results demonstrate that \modelname~is able to generate novel knowledge 
that humans rate as high quality, with up to 77.5\% (\textsc{Atomic}) and 91.7\% (ConceptNet) precision at top 1, which approaches human performance  
for these resources.
Our findings suggest that using generative commonsense models for automatic commonsense KB completion could soon be a plausible alternative to extractive methods.
\end{abstract}

\section{Introduction}
\label{sec:intro}

\begin{figure}[t]
    \centering
    \includegraphics[trim={5.75cm 7.25cm 9.8cm 1.25cm},clip,width=.48\textwidth]{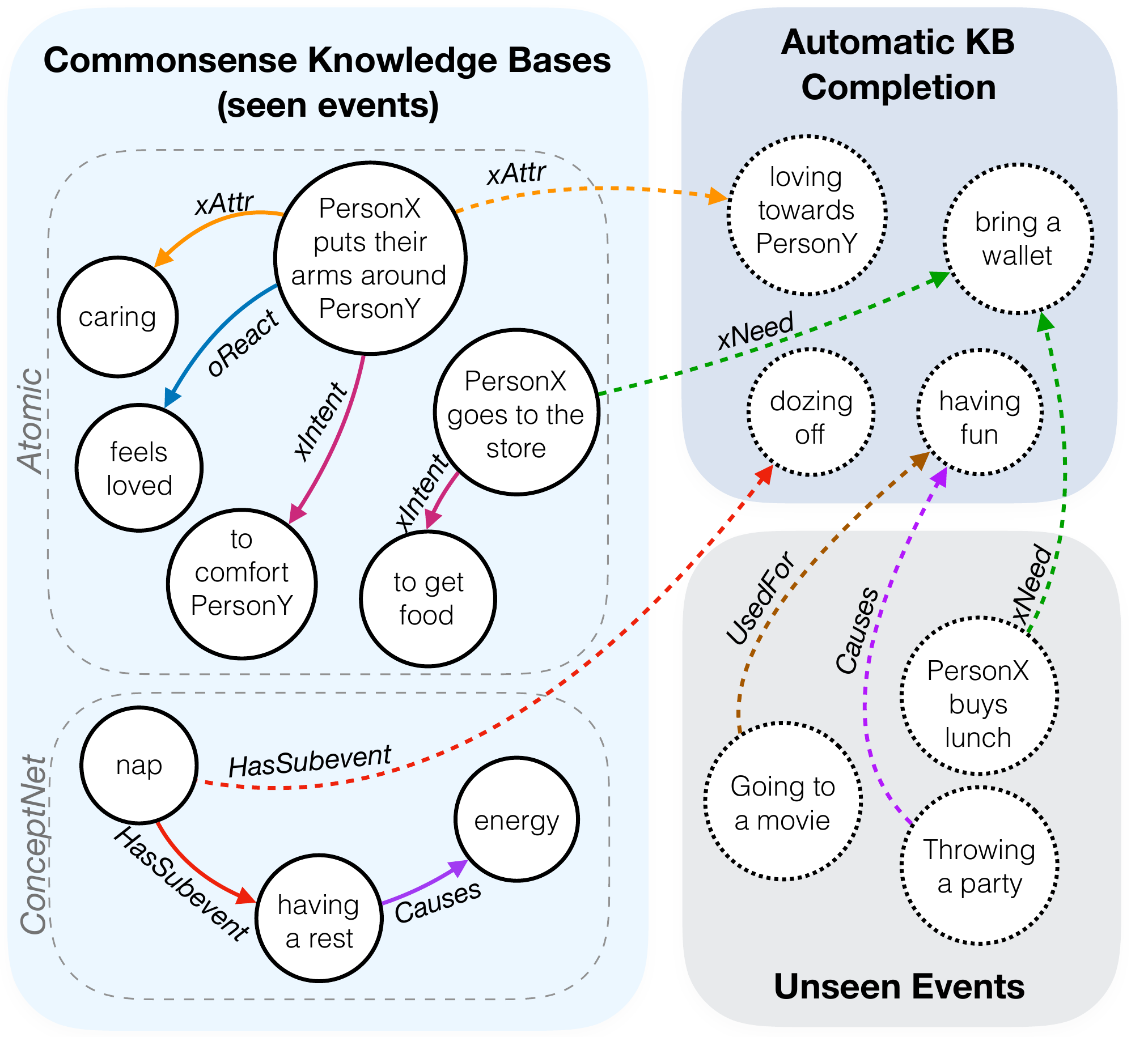}
    \caption{\modelname{}\cometemoji{} learns from an existing knowledge base (solid lines) to be able to generate novel nodes and edges (dashed lines).}
    \label{fig:intro}
\end{figure}

\begin{figure*}[t]
    \centering
    \includegraphics[trim={0cm 12cm 1.65cm 0cm},clip,width=.85\textwidth]{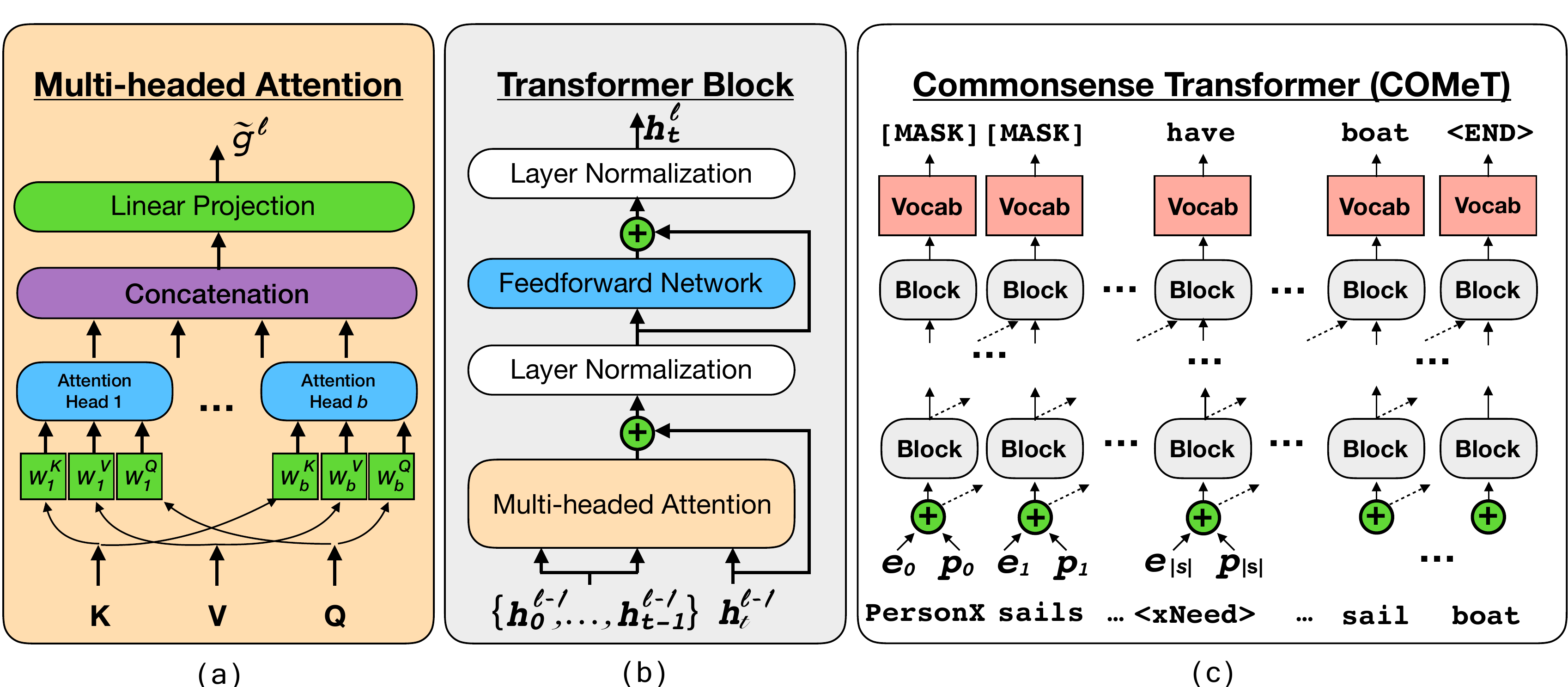}
    \caption{Model diagram. \textbf{(a)} In the multi-headed attention module, the key, value, and query all pass through a head-specific projection before a scaled dot-product attention is computed between them. The outputs of the heads are concatenated and projected. \textbf{(b)} Inside the transformer block, the outputs of all the previous layer blocks from earlier time steps are input to the multi-headed attention with the preceding block for the current time step as the query. \textbf{(c)} Each token is an input to a first-layer block along with all preceding tokens. Dotted lines indicate outputs to all future blocks in the next layer and inputs from all preceding blocks in the previous layer.}
    \label{fig:model}
\end{figure*}

When reading text, humans make commonsense inferences that frame their understanding of the narrative being presented. For machines to achieve this capability, they must be able to acquire relevant and correct commonsense for an unbounded set of situations. In this work, we cast commonsense acquisition as knowledge base construction and investigate whether large-scale language models can effectively learn to generate the knowledge necessary to automatically construct a commonsense knowledge base (KB).

Automatic KB construction is a long-standing goal of artificial intelligence research due to the difficulty of achieving high concept coverage in high-precision curated KBs \cite{lenat1995cyc,Miller1995}. Previous work has developed models capable of reading and extracting semi-structured text \cite{Suchanek2007Yago,Hoffart2013,Auer2007DBpediaAN,Bollacker2008FreeBase} and unstructured text \cite{Dong2014,Carlson2010,Nakashole2011,Nakashole2012,Niu2012} into relational schemas that can be queried for downstream applications. A common thread of these approaches, however, is the focus on encyclopedic 
knowledge, which lends itself to a well-defined space of entities and relations that can be modeled.

Commonsense knowledge, however, does not cleanly fit into a schema comparing two entities with a known relation, leading current approaches to model ``entities" as natural language phrases and relations as any concept that can link them \cite{li2016commonsense,sap2018atomic}. OpenIE approaches display this property of open text entities and relations \cite{Etzioni2011OpenIE,fader2011identifying,Mausam2012OpenLL}, but being extractive, they only capture knowledge that is explicitly mentioned in text, limiting their applicability for capturing commonsense knowledge, which is often implicit \cite{gordon2013reporting}.

Meanwhile, recent progress in training deep contextualized language models \cite{elmo, openaigpt, bert} provides an opportunity to explore beyond extractive methods as an avenue for commonsense KB construction. 
These large-scale language models display impressive performance when their underlying representations are tuned to solve end tasks, achieving state-of-the-art results on a variety of complex problems. In this work, we define the \emph{\textbf{COM}mons\textbf{E}nse \textbf{T}ransformer} (\modelname{}\cometemoji{}), which constructs commonsense KBs by using existing tuples as a seed set of knowledge on which to train. 
Using this seed set,
a pre-trained language model learns to adapt its learned representations to knowledge generation, and produces novel
tuples that are high quality.

We summarize our contributions in this work as follows. First, we develop a generative approach to knowledge base construction. A model must learn to produce new nodes and identify edges between existing nodes by generating phrases that coherently complete an existing seed phrase and relation type\footnote[1]{Demo is available at \url{https://mosaickg.apps.allenai.org/}}. Second, we develop a framework for using large-scale transformer language models to learn to produce commonsense knowledge tuples\footnote[2]{Code is available at \url{https://github.com/atcbosselut/comet-commonsense}}. Finally, we perform an empirical study on the quality, novelty, and diversity of the commonsense knowledge produced by our approach for two domains, \textsc{Atomic} and ConceptNet, as well as an efficiency study on the number of seed tuples needed to learn an effective knowledge model. The results indicate that \modelname~is able to produce high quality tuples as human judges find that 77.5\% of generated tuples for \textsc{Atomic} events and 91.7\% of generated tuples for ConceptNet relations are correct.

\section{Learning to Generate Commonsense}

\modelname~ is an adaptation framework for constructing commonsense knowledge bases from language models by training the language model on a seed set of knowledge tuples. These tuples provide \modelname~ with the KB structure and relations that must be learned, and \modelname~learns to adapt the language model representations learned from pre-training to add novel nodes and edges to the seed knowledge graph.

\subsection{Task}
\label{ssec:model:notation}
More specifically, the problem assumes \modelname~is given a training knowledge base of natural language tuples in 
$\{s, r, o\}$ format, 
where $s$ is the phrase subject of the tuple, $r$ is the relation of the tuple, and $o$ is the phrase object of the tuple.
For example, a ConceptNet tuple relating to ``taking a nap" would be: ($s$=``take a nap", $r$=\texttt{Causes}, $o$=``have energy"). The task is to generate $o$ given $s$ and $r$ as inputs.

\paragraph{Notation} We define $X^{s} = \{x^s_0, ..., x^s_{\vert s \vert}\}$ as the tokens that make up the subject of the relation, $X^{r} = \{x^r_0, ..., x^r_{\vert r \vert}\}$ as the tokens that make up the relation of the tuple, and $X^{o} = \{x^o_0, ..., x^o_{\vert o\vert}\}$ as the tokens that make up the object of the tuple. The embedding for any word $x$ is denoted as $e$.

\subsection{Transformer Language Model}
While \modelname~is agnostic to the language model with which it is initialized, in this work, we use the transformer language model architecture introduced in \citet{openaigpt} (GPT), which uses multiple transformer blocks of multi-headed scaled dot product attention and fully connected layers to encode input text \cite{Vaswani2017AttentionIA}. Figure~\ref{fig:model} depicts different components of the GPT architecture and we define each component in more depth below.

\paragraph{Transformer Block}

As shown in Figure~\ref{fig:model}(b), each transformer layer $l$ contains an architecturally identical transformer block (though with unique trainable parameters) that applies the following transformations to the input to the block:
\begin{align}
    \tilde g^l &= \textsc{MultiAttn}(h^{l-1}) \\
    g^{l} &= \textsc{LayerNorm}(\tilde g^l + h^{l-1}) \\
    \tilde h^l &= \textsc{FFN}(g^l) \\
    h^l &= \textsc{LayerNorm}(\tilde h^{l} + g^l)
\end{align}

\noindent where \textsc{MultiAttn} is a multi-headed self-attention mechanism (defined below), \textsc{FFN} is a two-layer feed-forward network, and \textsc{LayerNorm} represents a layer normalization \cite{layernorm} operation that is applied to the output of the self-attention and the feedforward network. Note that the inputs to the \textsc{LayerNorm} operations contain a residual connection that sums the output of and input to the previous operation.

\paragraph{Multi-headed Attention}

The multi-headed attention module of each transformer block, shown in Figure~\ref{fig:model}(a), is identical to the one originally defined by \citet{Vaswani2017AttentionIA}. The attention function receives three inputs, a query $Q$, key $K$, and value $V$. The attention is made of multiple $heads$ that each compute a unique scaled dot product attention distribution over $V$ using $Q$ and $K$:
\begin{align}
    \textsc{Attention}(Q, K, V) = \text{softmax}\bigg(\frac{QK^T}{\sqrt{d_k}}\bigg)V
\end{align}
\noindent where $d_k$ is the dimensionality of the input vectors representing the query, key and value. For each of the heads, $Q$, $K$, and $V$ are uniquely projected prior to the attention being computed:
\begin{align}
H_i &= \textsc{Attention}(QW^Q_i, KW^K_i, VW^V_i)
\end{align}
where $H_i$ is the output of a single attention head and $W^Q_i$, $W^K_i$, and $W^V_i$ are head-specific projections for $Q$, $K$, and $V$, respectively. The outputs of the attention heads $H_i$ are then concatenated:
\begin{align}
\textsc{MultiH(Q, K, V)} &= [H_1; ...; H_b]W^O
\end{align}
where $W^O$ is an output projection of the concatenated outputs of the attention heads. As shown in Figure~\ref{fig:model}(c), we follow \citet{openaigpt} and use the output of the previous layer's transformer block as the query input for the multi-headed attention of the next block. The keys and values are outputs of the previous layer's block for all preceding time steps:
\begin{align}
    \textsc{MultiAttn}(h_t^{l-1}) &=
    \textsc{MultiH}(h_t^{l-1},\mathbf{h}^{l-1}_t, \mathbf{h}^{l-1}_t)
\end{align}

\noindent where $\mathbf{h}^{l-1}_t = \{h^{l-1}\}_{<t}$ is the set of previous layer transformer block outputs for time steps preceding $t$.

\paragraph{Input Encoder} As input to the model, we represent a knowledge tuple $\{s, r, o\}$ as a concatenated sequence of the words of each item of the tuple:
\begin{equation}
\mathbf{X} = \{X^{s}, X^{r}, X^{o}\}
\label{eq:sequence}
\end{equation}

\noindent Since the transformer (a self-attention model) has no concept of ordering of tokens, a position embedding $p_t$ is initialized for each absolute position in the sequence \citep{Vaswani2017AttentionIA}. For any input word $x_t \in \mathbf{X}$, our encoding of the input is the sum of its word embedding, $e_t$ with a position embedding encoding its absolute position in the sequence $\mathbf{X}$:

\begin{equation}
h_t^0 = e_t + p_t
\label{eq:input}
\end{equation}

\noindent where $p_t$ is the position embedding for time step $t$, and $h^0$ is the input to the first transformer layer.

\begin{figure}[t]
    \centering
    \includegraphics[trim={2cm 13.0cm 2cm 2.25cm},clip,width=\columnwidth]{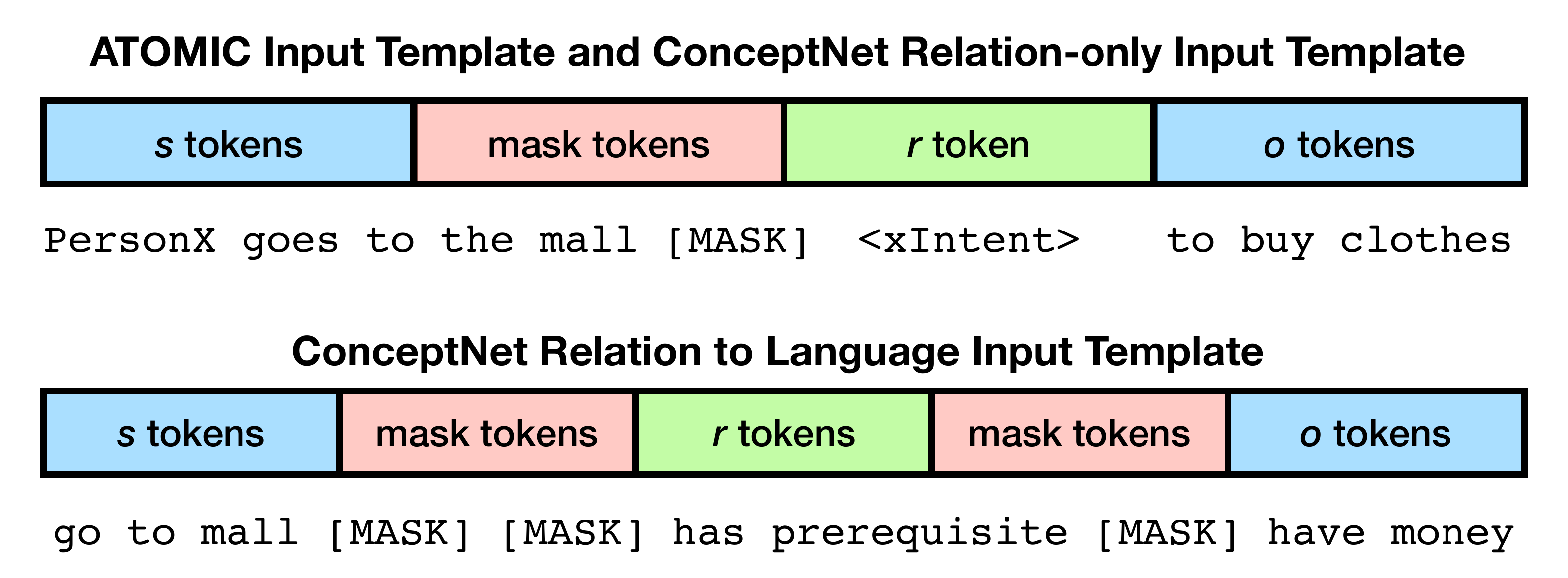}
    \caption{Input token setup for training configurations. For the \textsc{Atomic} dataset, the tokens of the subject, $X^s$ (e.g., PersonX goes to the mall) are followed by masking tokens, which is followed by a single relation token $X^r$ (e.g., \texttt{xIntent}), and then the object tokens $X^o$ (e.g., to buy clothes). The model receives the same input for ConceptNet, except that a second set of masking tokens separate $X^r$ and $X^o$ because $X^r$ can have a variable number of tokens for ConceptNet (\S\ref{ssec:conceptnet:results})}
    \label{fig:tokens}
\end{figure}
\section{Training \modelname}
\label{sec:train}

\modelname~is trained to learn to produce the phrase object $o$ of a knowledge tuple given the tuple's phrase subject $s$ and relation $r$. More specifically, given the concatenation of the tokens of $s$ and $r$: $[X^s, X^r]$ as input, the model must learn to generate the tokens of $o$: $X^o$ (See \S\ref{ssec:model:notation} for definitions of these variables).

\paragraph{Loss Function}
\label{ssec:train:loss} To achieve this goal, \modelname~is trained to maximize the conditional loglikelihood of predicting the phrase object tokens, $X^o$:

\begin{equation}
\mathcal{L} = - \sum_{t= \vert s \vert + \vert r \vert }^{\vert s \vert + \vert r \vert + \vert o \vert} \log P(x_t | x_{<t})
\label{eq:loss}
\end{equation}

\noindent where $\vert s \vert$, $\vert r \vert$, and $\vert o \vert$ are the number of tokens in the subject phrase, relation, and object phrase, respectively. Figure~\ref{fig:tokens} outlines how the tokens in $s$, $r$, and $o$ are organized for different training tasks.

\paragraph{Datasets}
\label{ssec:train:datasets}

\modelname~relies on a seed set of knowledge tuples from an existing KB to learn to produce commonsense knowledge.
In this work, we use \textsc{Atomic} and ConceptNet as knowledge seed sets, but other commonsense knowledge resources could have been used as well as \modelname~is domain-agnostic.

\begin{table*}[t]
\centering
\footnotesize
\resizebox{0.8\linewidth}{!}{
\begin{tabular}{l rrr  rrr  rrr r}
\toprule
 \textbf{Model} & \textbf{PPL}\footnotemark[5] & \textbf{BLEU-2} &  \textbf{N/T $sro$}\footnotemark[6] & \textbf{N/T $o$} & \textbf{N/U $o$}  \\
 \toprule
 \textsc{9Enc9Dec} \cite{sap2018atomic} & - & 10.01  & 100.00         & 8.61 & 40.77  \\
 NearestNeighbor \cite{sap2018atomic} & - & 6.61 & -             & -    & -      \\
 Event2\textsc{(In)Volun} \cite{sap2018atomic} & - & 9.67 & 100.00    & 9.52 & 45.06    \\
  Event2\textsc{PersonX/Y} \cite{sap2018atomic} & - & 9.24 & 100.00   & 8.22 & 41.66   \\
 Event2\textsc{Pre/Post} \cite{sap2018atomic} & - & 9.93 & 100.00     & 7.38 & 41.99    \\     
\midrule
\modelname~ (- pretrain) & 15.42 & 13.88 & 100.00      & 7.25& 45.71  \\
 \modelname & \textbf{11.14} & \textbf{15.10} &  100.00                      & \textbf{9.71} & \textbf{51.20}   \\

\bottomrule
\end{tabular}}
\caption{Automatic evaluations of quality and novelty for generations of \textsc{Atomic} commonsense. No novelty scores are reported for the NearestNeighbor baseline because all retrieved sequences are in the training set.}
\label{tab:atomic:auto}
\end{table*}

\definecolor{lightgray}{rgb}{0.95, 0.95, 0.95}
\newcolumntype{g}{>{\columncolor{lightgray}}c}

\begin{table*}[t]
\centering
\resizebox{\linewidth}{!}{
\begin{tabular}{l || rrr  rrr  rrr ||g}
\toprule
 \textbf{Model} & \textbf{oEffect} & \textbf{oReact} & \textbf{oWant} & \textbf{xAttr} & \textbf{xEffect} & \textbf{xIntent} & \textbf{xNeed} & \textbf{xReact} & \textbf{xWant} & \textbf{Avg} \\ 
 \toprule
9Enc9Dec \cite{sap2018atomic} & 22.92 & 32.92 & 35.50 & 52.20 & 47.52 & 51.70 & 48.74 & 63.57 & 51.56 & 45.32 \\
Event2(In)voluntary \cite{sap2018atomic} & \underline{26.46} & 36.04 & 34.70 & 52.58 & 46.76 & 61.32 & 49.82 & 71.22 & 52.44 & 47.93 \\
Event2PersonX/Y \cite{sap2018atomic} & 24.72 & 33.80 & 35.08 & \underline{52.98} & 48.86 & 53.93 & 54.05 & 66.42 & 54.04 & 46.41 \\
Event2Pre/Post \cite{sap2018atomic} & \underline{26.26} & 34.48 & 35.78 & 52.20 & 46.78 & 57.77 & 47.94 & 72.22 & 47.94 & 46.76 \\
\midrule
 \modelname~(- pretrain) & \underline{25.90} & \underline{35.40} & \underline{40.76} & 48.04 & 47.20 & 58.88 & 59.16 & 64.52  & 65.66 & 49.50 \\
\modelname & \textbf{29.02} & \textbf{37.68} & \textbf{44.48} & \textbf{57.48} & \textbf{55.50} & \textbf{68.32} & \textbf{64.24} & \textbf{76.18} & \textbf{75.16} & \textbf{56.45} \\
\bottomrule
\end{tabular}}
\caption{Human score of generations of \textsc{Atomic} commonsense. We present comparisons to the baselines from \citet{sap2018atomic}.
\underline{Underlined} results are those where \modelname{} is \underline{not} significantly better at $p<0.05$ }
\label{tab:atomic:human}
\end{table*}

\paragraph{Initialization} 
Parameters are initialized to the final language model weights from \citet{openaigpt}.
Additional special tokens that are added to the vocabulary for fine tuning (e.g., relation embeddings such as \texttt{oReact} for \textsc{Atomic} and \texttt{IsA} for ConceptNet) are initialized by sampling from the standard normal distribution. 

\paragraph{Hyperparameters} Following \citet{openaigpt}'s  design of the GPT model, we initialize  \modelname~with 12 layers, 768-dimensional hidden states, and 12 attention heads. We use a dropout rate of 0.1 and use GeLU \cite{gelu} units as activation functions. During training, our batch size is 64.
Other dataset-specific hyperparameters are provided in Appendix~\ref{app:train:hyper}.

\section{\textsc{Atomic} Experiments}
\label{sec:atomic}

The \textsc{Atomic} dataset\footnote[3]{\url{https://homes.cs.washington.edu/~msap/atomic/}}, released by \citet{sap2018atomic}, contains 877K tuples covering a variety of social commonsense knowledge around specific event prompts (e.g., ``X goes to the store'').
Specifically, \textsc{Atomic} distills its commonsense in nine dimensions, covering the event's causes (
e.g., ``X needs to drive there''), its effects on the agent (
e.g., ``to get food'') and its effect on other direct (or implied) participants (
e.g., ``Others will be fed''). 
More details about \textsc{Atomic} can be found in Appendix~\ref{app:atomic:train}.
For our experiments, \textsc{Atomic} events (e.g., ``X goes to the store'') are phrase subjects, $s$, the dimension (e.g., \texttt{xIntent}) is the phrase relation, $r$, and the causes/effects (e.g., ``to get food'') are phrase objects, $o$.
We use the training splits from~\citet{sap2018atomic}, resulting in 710k training, 80k development, and 87k test tuples respectively.

\subsection{Setup}
\label{ssec:atomic:setup}
\paragraph{Metrics}

Following \citet{sap2018atomic}, we evaluate our method using BLEU-2 as an automatic evaluation metric. We also report the perplexity of the model on its gold generations. The remaining automatic metrics in Table~\ref{tab:atomic:auto} measure the proportion of generated tuples and generated objects which are not in the training set. We report the proportion of all generated tuples that are novel (\% N/T $sro$) and that have a novel object (\% N/T $o$)\footnote[4]{a new $o$ represents a new node in the knowledge graph}. To show that these novel objects are diverse (i.e., the same novel object is not the only one being generated), we also report the number of novel objects as a function of the set of \textit{unique} objects produced for all test set events (\% N/U $o$). 

Finally, we perform a human evaluation using workers from Amazon Mechanical Turk (AMT). Workers are asked to identify whether a model generation of \textsc{Atomic} commonsense adequately completes a plausible tuple of phrase subject, relation, and phrase object. Following the setup of \citet{sap2018atomic}, we evaluate 100 randomly selected events from the test set. For each event and relation type, 10 candidates are generated using beam search and the full beam is evaluated by five different workers. Overall, n=5000 ratings are produced per relation (100 events $\times$ 5 workers $\times$ 10 candidates). The reported \textbf{Avg} in Table~\ref{tab:atomic:human} is an average of these scores, yielding n=45000 total ratings for each model. We use Pitman's test \cite{Pitman} with 100k permutations to test for statistical significance. Because 50 different hypotheses are tested (9 relations + the total), the Holm-Bonferroni method \cite{Holm1979} is used to correct significance thresholds. Example events from the development set and their generated phrase objects are available in Table~\ref{tab:atomic_generations}.

\paragraph{Baselines}

We report the performance of our method against the models trained in \citet{sap2018atomic} that use LSTM sequence-to-sequence models \cite{seq2seq} to encode the input subject and relation and produce an output object. 

\paragraph{Ablations} 
To evaluate how pre-training on a large corpus helps the model learn to produce knowledge, we train a version of \modelname~that is not initialized with pre-trained weights (\modelname~(- pretrain)).
We also evaluate the data efficiency of our method by training models on different proportions of the training data.
Finally, because the ultimate goal of our method is to be able to perform high-quality, diverse knowledge base construction, we explore how various decoding schemes affect the quality of candidate knowledge tuples. We present the effect of the following generation strategies: argmax greedy decoding, beam search with beam sizes, b=2, 5, 10, and top-$k$ sampling with k = 5, 10. For each decoding method, we conduct the human evaluation on the number of final candidates produced by each method.

\footnotetext[5]{\citet{sap2018atomic}'s models were trained with a different vocabulary so a direct perplexity comparison is not possible.}
\footnotetext[6]{All test set $s$ do not appear in the training set so all full tuples must be novel.}

\begin{table*}[t]
\centering
\resizebox{\linewidth}{!}{

\begin{tabular}{l || rrr  rrr  rrr ||g}

\multicolumn{9}{c}{ }\\
\toprule
 \modelname~ \textbf{Decoding method} & \textbf{oEffect} & \textbf{oReact} & \textbf{oWant} & \textbf{xAttr} & \textbf{xEffect} & \textbf{xIntent} & \textbf{xNeed} & \textbf{xReact} & \textbf{xWant} & \textbf{Avg} \\ 
\midrule
Top-5 random sampling (n=2500 per relation) & 34.60 &  44.04 &  35.56 &  64.56 &  55.68 &  58.84 &  46.68 &  80.96 &  58.52 &  53.27 \\
Top-10 random sampling (n=5000 per relation) & 25.20 &  37.42 &  27.34 &  49.20 &  47.34 &  47.06 &  38.24 &  72.60 &  48.10 &  43.61 \\
Beam search - 2 beams (n=1000 per relation) & 43.70 &  54.20 &  47.60 &  \textbf{84.00} &  51.10 &  73.80 &  50.70 &  85.80 &  78.70 &  63.29 \\
Beam search - 5 beams (n=2500 per relation) & 37.12 &  45.36 &  42.04 &  63.64 &  \textbf{61.76} &  63.60 &  57.60 &  78.64 &  68.40 &  57.57 \\
Beam search - 10 beams (n=5000 per relation) & 29.02 &  37.68 &  44.48 &  57.48 &  55.50 &  68.32 &  64.24 &  76.18 &  75.16 &  56.45 \\
Greedy decoding (n=500 per relation) & \textbf{61.20} &  \textbf{69.80} &  \textbf{80.00} &  77.00 &  53.00 &  \textbf{89.60} &  \textbf{85.60} &  \textbf{92.20} &  \textbf{89.40} &  \textbf{77.53} \\
\midrule
Human validation of gold \textsc{Atomic} & 84.62 & 86.13 & 83.12 & 78.44 & 83.92 & 91.37 & 81.98 & 95.18 & 90.90 & 86.18 \\
\bottomrule
\end{tabular}}
\caption{Human evaluation testing effect of different decoding schemes on candidate tuple quality. The number of ratings made per relation for each decoding method is provided in the first column.}
\label{tab:atomic:human:decoding}
\end{table*}
\subsection{Results} 
\label{ssec:atomic:results}
\paragraph{Overall performance}

The BLEU-2 results in Table~\ref{tab:atomic:auto} indicate that \modelname~exceeds the performance of all baselines, achieving a 51\% relative improvement over the top performing model of \citet{sap2018atomic}. More interesting, however, is the result of the human evaluation, where \modelname~ reported a statistically significant relative \textbf{Avg} performance increase of 18\% over the top baseline, Event2\textsc{In(Volun)}. This performance increase is consistent, as well, with an improvement being observed across every relation type. In addition to the quality improvements, Table~\ref{tab:atomic:auto} shows that \modelname{} produces more novel tuple objects than the baselines, as well.

\begin{table}[t]
\centering
\small
\begin{tabular}{r  rrrr}
\toprule
 \textbf{\% train data} & \textbf{PPL} & \textbf{BLEU-2} & \textbf{N/T $o$} & 
 \textbf{N/U $o$}   \\
\midrule 
1\% train  & 23.81 & 5.08  & 7.24 & 49.36\\
10\% train & 13.74 & 12.72  & \textbf{9.54} & \textbf{58.34} \\
50\% train & 11.82 & 13.97 & 9.32 & 50.37 \\
\midrule
\textsc{Full} (- pretrain) & 15.18 & 13.22  & 7.14 & 44.55\\
\midrule
\textsc{Full} train & \textbf{11.13} & \textbf{14.34} & 9.51 & 50.05 \\
\bottomrule
\end{tabular}
\caption{Effect of amount of training data on automatic evaluation of commonsense generations}
\label{tab:atomic:efficiency}
\end{table}

\paragraph{Learning knowledge from language} Significant differences were also observed between the performance of the model whose weights were initialized with the pre-trained parameters from the GPT model of \citet{openaigpt} and a model with the same architecture that was trained from random initialization. This 14\% relative improvement in overall human performance confirms that the language representations learned by the GPT model are transferable to generating natural language commonsense knowledge.

\paragraph{Effect of decoding algorithm} In  Table~\ref{tab:atomic:human:decoding},  we show the effect of different generation policies on knowledge quality. The most interesting result is that using greedy decoding to produce knowledge tuples only results in a 10\% relative performance gap compared to a human evaluation of the \textsc{Atomic} test set, showing that the knowledge produced by the model approaches human performance. While producing more total candidates does lower overall performance, quality assessments still hover around 55\%\footnote[7]{This number is partially low due to the many ``none" references in the \texttt{oEffect}, \texttt{oReact}, \texttt{oWant} categories. In any set of 10 candidates, ``none" can only be predicted once, which causes most candidates in the beam to be incorrect if ``none" is the appropriate answer.} for a beam size of 10. This result suggests that \modelname~could be effective with human evaluators in the loop to confirm the correctness of generated tuples.

\begin{table}[t!]
\centering
\scriptsize
\resizebox{\columnwidth}{!}{
\begin{tabular}{l l l c}
\toprule
 \multicolumn{1}{l}{\textbf{Seed Concept}} & \multicolumn{1}{l}{\textbf{Relation}} & \multicolumn{1}{l}{\textbf{Generated}} &  \multicolumn{1}{l}{\textbf{Plausible}}\\ 
\toprule
\rowcolor{Gray}
 X holds out  X's hand to  Y  & \texttt{xAttr} & helpful & \checkmark \\
\rowcolor{Gray}
 X meets  Y eyes  & \texttt{xAttr} & intense & \checkmark \\
\rowcolor{Gray}
 X watches  Y every \texttt{\_\_\_}  & \texttt{xAttr} & observant & \checkmark\\
 X eats red meat  & \texttt{xEffect} & gets fat & \checkmark\\
 X makes crafts  & \texttt{xEffect} & gets dirty & \checkmark\\
 X turns  X's phone  & \texttt{xEffect} & gets a text & \\ 
\rowcolor{Gray}
 X pours \texttt{\_\_\_} over  Y's head  & \texttt{oEffect} & gets hurt & \checkmark\\
\rowcolor{Gray}
 X takes  Y's head off  & \texttt{oEffect} & bleeds & \checkmark\\
\rowcolor{Gray}
 X pisses on  Y's bonfire  & \texttt{oEffect} & gets burned & \\
 X spoils somebody rotten  & \texttt{xIntent} & to be mean &  \\ 
 X gives  Y some pills  & \texttt{xIntent} & to help & \checkmark\\
 X provides for  Y's needs  & \texttt{xIntent} & to be helpful & \checkmark\\
\rowcolor{Gray}
 X explains  Y's reasons  & \texttt{xNeed} & to know  Y & \checkmark\\
\rowcolor{Gray}
 X fulfils  X's needs  & \texttt{xNeed} & to have a plan & \checkmark\\
\rowcolor{Gray}
 X gives  Y everything  & \texttt{xNeed} & to buy something & \checkmark\\
 X eats pancakes  & \texttt{xReact} & satisfied & \checkmark\\
 X makes \texttt{\_\_\_} at work  & \texttt{xReact} & proud & \checkmark\\
 X moves house  & \texttt{xReact} & happy & \checkmark\\
\rowcolor{Gray}
 X gives birth to the  Y  & \texttt{oReact} & happy & \checkmark\\
\rowcolor{Gray}
 X gives  Y's friend \texttt{\_\_\_}  & \texttt{oReact} & grateful & \checkmark\\
\rowcolor{Gray}
 X goes \texttt{\_\_\_} with friends  & \texttt{oReact} & happy & \checkmark\\
 X gets all the supplies  & \texttt{xWant} & to make a list & \checkmark\\
 X murders  Y's wife  & \texttt{xWant} & to hide the body & \checkmark\\
 X starts shopping  & \texttt{xWant} & to go home & \checkmark\\
\rowcolor{Gray}
 X develops  Y theory  & \texttt{oWant} & to thank  X & \checkmark\\
\rowcolor{Gray}
 X offer  Y a position  & \texttt{oWant} & to accept the job & \checkmark\\
\rowcolor{Gray}
 X takes \texttt{\_\_\_} out for dinner  & \texttt{oWant} & to eat & \checkmark\\
\bottomrule
\end{tabular}
}
\caption{Generations that were \textbf{randomly selected} from a subset of \textbf{novel} generations from the \textsc{Atomic} development set. A novel generation is a $sro$ tuple not found in the training set. Manual evaluation of each tuple indicates whether the tuple is considered plausible by a human annotator.}
\label{tab:atomic_generations}
\end{table}

\paragraph{Efficiency of learning from seed tuples} Because not all domains will have large available commonsense KBs on which to train, we explore how varying the amount of training data available for learning affects the quality and novelty of the knowledge that is produced. Our results in Table~\ref{tab:atomic:efficiency} indicate that even with only 10\% of the available training data, the model is still able to produce generations that are coherent, adequate, and novel. Using only 1\% of the training data clearly diminishes the quality of the produced generations, with significantly lower observed results across both quality and novelty metrics. Interestingly, we note that training the model without pre-trained weights performs comparably to training with 10\% of the seed tuples, quantifying the impact of using pre-trained language representations.

\section{ConceptNet Experiments}
\label{sec:conceptnet}

The ConceptNet dataset\footnote[8]{\url{https://ttic.uchicago.edu/~kgimpel/commonsense.html}}, provided by \citet{li2016commonsense}, consists of tuples obtained from the Open Mind Common Sense (OMCS) entries in ConceptNet 5 \cite{speer2017conceptnet}. Tuples are in the standard $sro$ form -- (e.g., take a nap, \texttt{Causes}, have energy). The most confident 1200 tuples were used to create the test set, while the next 1200 tuples were used to create two development sets, which we combine in this work. The 100k version of the training set was used to train models, which contains 34 relation types.

\subsection{Setup}
\label{ssec:conceptnet:setup}
\paragraph{Metrics}

We evaluate our models that generate ConceptNet relations using the following metrics. First, we report the perplexity of the gold relations in the test set (PPL). 
To evaluate the quality of generated knowledge, we also report the number of generated positive examples in the test set that are scored as correct by the pre-trained Bilinear AVG model developed by 
\citet{li2016commonsense}.\footnote[9]{ A pre-trained model can be found at \url{https://ttic.uchicago.edu/~kgimpel/comsense_resources/ckbc-demo.tar.gz}} For a given $sro$ tuple, this model produces a probability for whether the tuple is correct. We threshold scores at 50\% probability to identify positive predictions. On the completion task originally proposed in \citet{li2016commonsense}, this model achieved 92.5\% accuracy on the test set, indicating that it is a strong proxy for automatically evaluating whether a generated tuple is correct. 
Finally, we report the same novelty metrics as for  \textsc{Atomic}: \textbf{N/T} $sro$ and \textbf{N/T} $o$.

\paragraph{Baselines} As a baseline, we re-implement the BiLSTM model proposed by \citet{saito2018commonsense} with minor modifications outlined in Appendix~\ref{app:train:baseline}. This model is trained to learn to encode knowledge in both directions: $sr \rightarrow o$ and $or \rightarrow s$ to help augment a knowledge base completion model. It is only evaluated on the $sr \rightarrow o$ tuple generation task, however. For posterity, we also include the result from a LSTM model that is only trained on the $sr \rightarrow o$ task (LSTM - $s$).

\paragraph{Ablations} We include the following ablations of our full model. First, we evaluate how pre-training on a large-scale corpus \cite{openaigpt} helps performance by training a comparison model from scratch, denoted \modelname~(- pretrain) in Table~\ref{tab:conceptnet}. Second, in our main model, we map relation names to natural language (e.g., \texttt{IsA} $\to$ ``is a''; \texttt{HasSubevent} $\to$ ``has subevent'') so the model can learn to represent these concepts with language, as opposed to learning a special embedding from scratch for each relation \cite{Levy2017ZeroShotRE}. As an ablation, we train a model without converting relation tokens to natural language (e.g., \texttt{IsA} $\not\to$ ``is a''), which we denote \modelname~- \textsc{RelTok}. 

\begin{table}[t]
\centering
\resizebox{\linewidth}{!}{
\begin{tabular}{l  rrr  rrrr}
\toprule
 \textbf{Model}
          & \textbf{PPL} & \textbf{Score} & \textbf{N/T $sro$} & \textbf{N/T $o$} & \textbf{Human} \\
\toprule 
LSTM - $s$ & - & 60.83 & \textbf{86.25} & 7.83 & 63.86 \\
CKBG \cite{saito2018commonsense} & - & 57.17 & \textbf{86.25} & \textbf{8.67} & 53.95\\
\midrule
\modelname~(- pretrain) & 8.05          & 89.25         & 36.17         & 6.00 & 83.49  \\
\modelname~- \textsc{RelTok}    & 4.39          & 95.17         & 56.42         & 2.62          & \textbf{92.11} \\
\midrule
\modelname~ & \textbf{4.32} & \textbf{95.25}& 59.25 & 3.75          & 91.69 \\
\bottomrule
\end{tabular}
}
\caption{ConceptNet generation Results}
\label{tab:conceptnet}
\end{table}

\subsection{Results}
\label{ssec:conceptnet:results}
\paragraph{Quality} Our results indicate that high-quality knowledge can be generated by the model: the low perplexity scores in Table~\ref{tab:conceptnet} indicate high model confidence in its predictions, while the high classifier score (95.25\%) indicates that the KB completion model of \citet{li2016commonsense} scores the generated tuples as correct in most of the cases. While adversarial generations could be responsible for this high score, a human evaluation (following the same design as for \textsc{Atomic}) scores 91.7\% of greedily decoded tuples as correct. 
Randomly selected examples provided in Table~\ref{tab:cn_generations} also point to the quality of knowledge produced by the model.

\paragraph{Novelty} In addition to being high quality, the generated tuples from \modelname~are also novel, with 59.25\% of the tuples not being present in the training set, showing that the model is capable of generating new edges between nodes, and even creating new nodes -- 3.75\% of $o$ nodes are novel -- to extend the size of the knowledge graph. One shortcoming, however, is that novel generations are sometimes simplified forms of tuples from the training set. In Table~\ref{tab:cn_generations}, for example, the tuple ``doctor \texttt{CapableOf} save life'' is not present in the training set, but ``doctor \texttt{CapableOf} save person life'' is. Many tuples, however, are completely novel, such as ``bird bone \texttt{HasProperty} fragile'' and ``driftwood \texttt{AtLocation} beach'', which have no related tuples in the training set.  

To explore further, we investigate by how much novel tuples from the development set differ from training set phrase objects for the same $s,r$ using minimum edit distance of phrase objects.
\begin{figure}
    \centering
    \includegraphics[trim={9.75cm 8.25cm 10.1cm 9cm}, clip,width=.98\columnwidth]{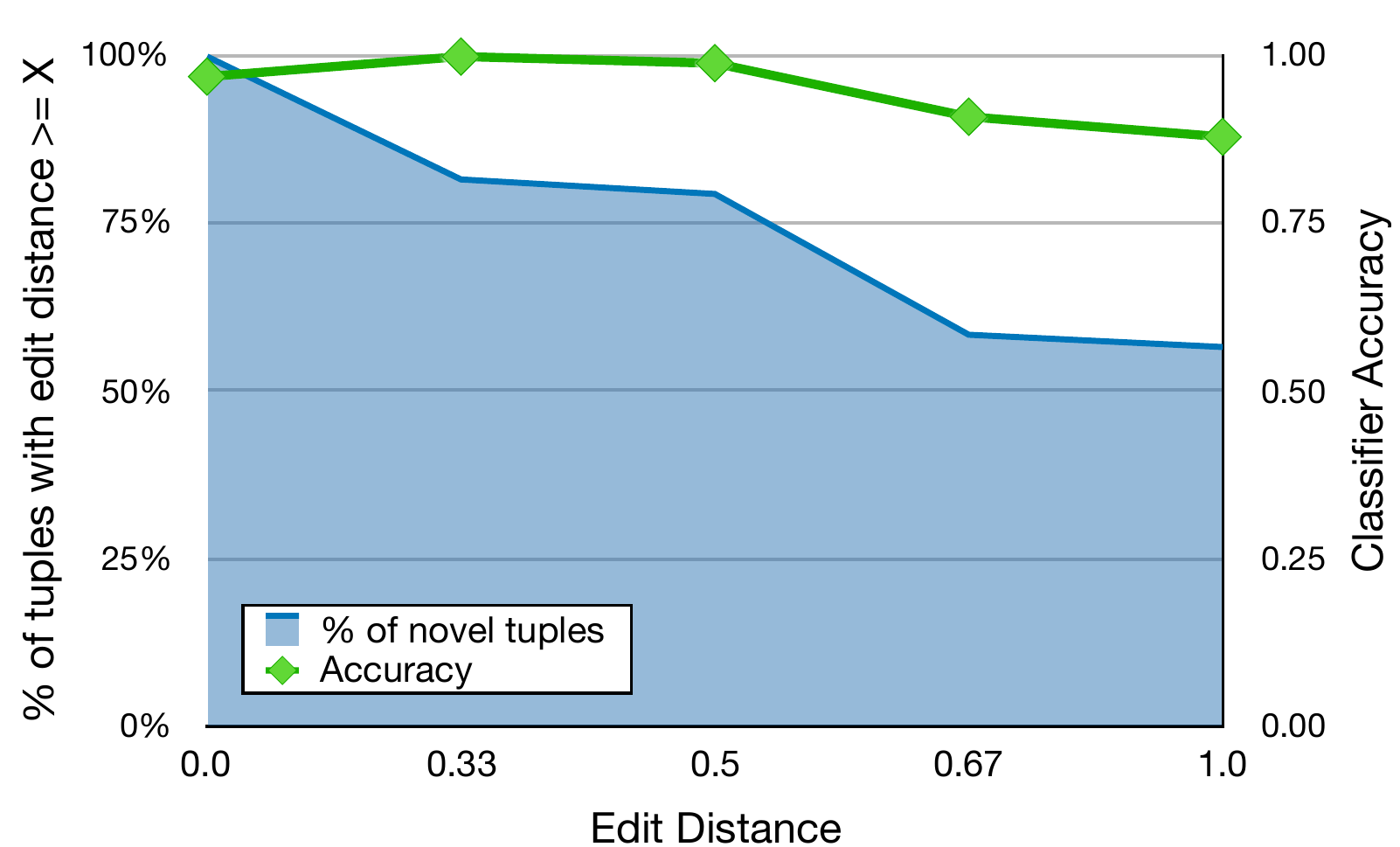}
    \caption{The percentage of novel ConceptNet development set tuples per minimum edit distance from training tuples.  In green: classifier-scored accuracy of each subset.}
    \label{fig:editdistcurve}
\end{figure}
We measure the edit distance of phrase object $o_{dev}$ in the tuple $(s,r,o_{dev})$ to the $o_{trn}$ from the nearest training tuple $(s,r,o_{trn})$.  Edit distance is measured using word tokens (excluding stop words) and normalized by the maximum number of words in $o_{dev}$ or $o_{trn}$.  The maximum edit distance is one (i.e., entirely different word sequences) and the minimum edit distance is zero (i.e., the same sequence excluding stopwords).
Figure~\ref{fig:editdistcurve} shows the percentage of novel development set tuples that have an edit distance from the closest training set tuple of at least the value on the x-axis. Over 75\% of the novel tuples have objects that are a normalized edit distance of $>=0.5$ from the training phrase objects, indicating that most of the novel phrase objects have significantly different word sequences from their closest analogues in the training set.

\paragraph{Learning knowledge from language} Similarly to \textsc{Atomic}, we explore how pre-training \modelname~ on a large language corpus affects its ability to generalize commonsense. This effect is apparent in Table~\ref{tab:conceptnet}, with a clear improvement on automatic and human evaluations by the pretrained \modelname~over the randomly initialized model. Qualitatively, we observe this effect in Table~\ref{tab:cn_generations} with the generated example tuple ``mango \texttt{IsA} fruit", which is not present in the training set. The only tuple containing the ``mango" entity in the training set is ``mango \texttt{UsedFor} salsa", which is not informative enough. As confirmation, we observe that the output from \modelname~(- pretrain) is ``mango \texttt{IsA} spice'', which could be a reasonable inference given the information about ``mango" in the seed set of knowledge.

\paragraph{Representing relations with language} While the automatic metrics point to insignificant differences when comparing models with symbol relations and those with natural language relations (Table~\ref{tab:conceptnet}), examples can provide qualitative insights into the benefits of representing relations as language. While the only non-ornithological reference to a ``dove" in the ConceptNet training set is ``dove \texttt{CapableOf} fly'', our model learns to generalize to produce the tuple ``dove \texttt{SymbolOf} purity''. The model that uses symbol relation embeddings only manages to produce the relation ``dove \texttt{SymbolOf} submarine'', which seems to relate ``submarine" to a more nautical (and unrelated) word sense of ``dove".

\begin{table}[t!]
\centering
\scriptsize
\resizebox{\linewidth}{!}{
\begin{tabular}{l l l c}
\toprule
 \multicolumn{1}{l}{\textbf{Seed}} & \multicolumn{1}{l}{\textbf{Relation}} & \multicolumn{1}{l}{\textbf{Completion}} &  \multicolumn{1}{l}{\textbf{Plausible}}\\ 
\toprule
piece & \texttt{PartOf} & machine & \checkmark \\
\rowcolor{Gray}
bread & \texttt{IsA} & food & \checkmark \\
\rowcolor{Gray}
oldsmobile & \texttt{IsA} & car & \checkmark \\
\rowcolor{Gray}
happiness & \texttt{IsA} & feel & \checkmark \\
\rowcolor{Gray}
math & \texttt{IsA} & subject & \checkmark \\
\rowcolor{Gray}
mango & \texttt{IsA} & fruit & \checkmark \\
\rowcolor{Gray}
maine & \texttt{IsA} & state & \checkmark \\
planet & \texttt{AtLocation} & space & \checkmark \\
dust & \texttt{AtLocation} & fridge & \\
puzzle & \texttt{AtLocation} & your mind & \includegraphics[width=1em]{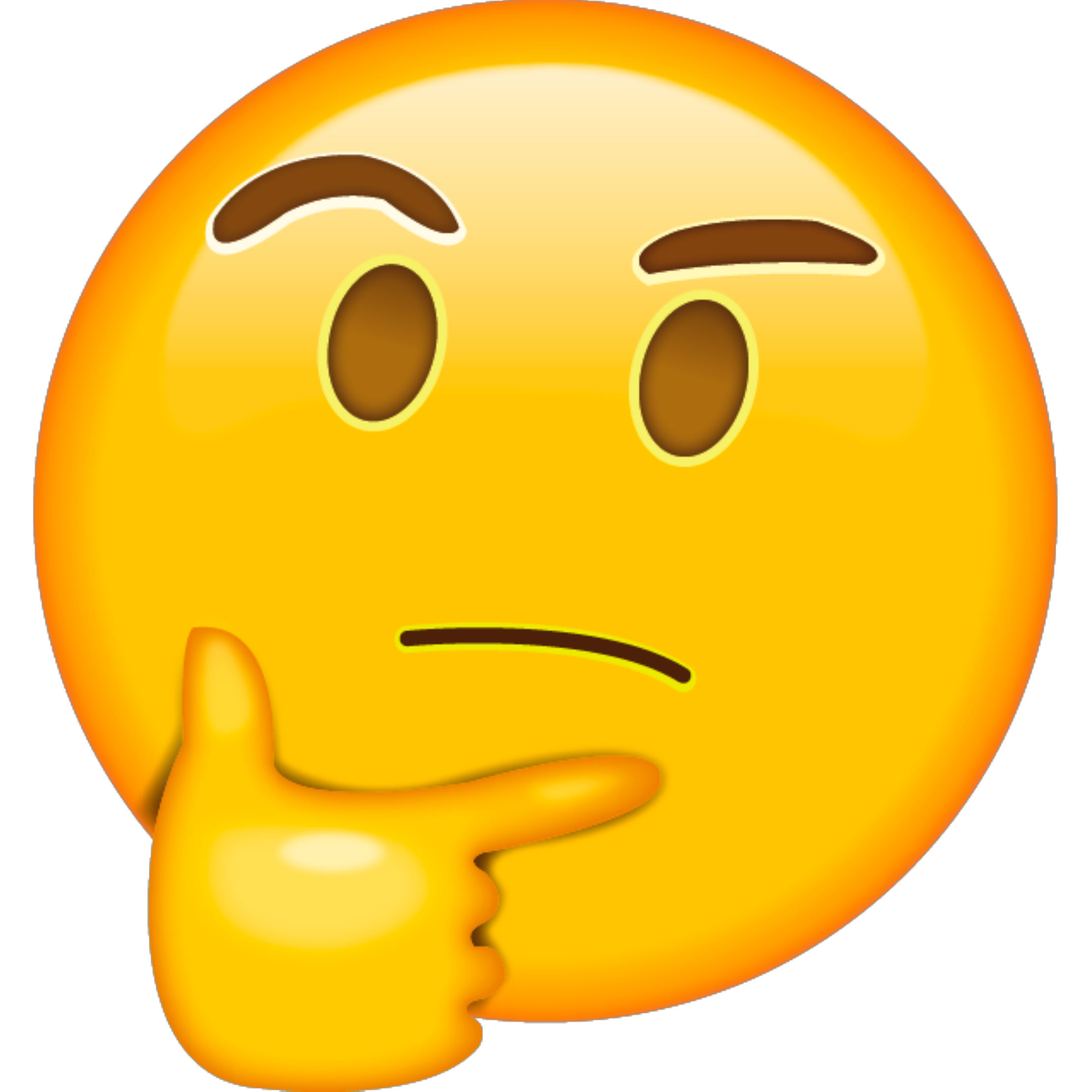}\\
college & \texttt{AtLocation} & town & \checkmark \\
dental chair & \texttt{AtLocation} & dentist & \checkmark \\
finger & \texttt{AtLocation} & your finger & \\
\rowcolor{Gray}
sing & \texttt{Causes} & you feel good & \checkmark \\
doctor & \texttt{CapableOf} & save life & \checkmark \\
post office & \texttt{CapableOf} & receive letter & \checkmark \\
\rowcolor{Gray}
dove & \texttt{SymbolOf} & purity & \checkmark \\
sun & \texttt{HasProperty} & big & \checkmark \\
bird bone & \texttt{HasProperty} & fragile & \checkmark \\
\rowcolor{Gray}
earth & \texttt{HasA} & many plant & \checkmark \\
yard & \texttt{UsedFor} & play game & \checkmark \\
\rowcolor{Gray}
get pay & \texttt{HasPrerequisite} & work & \checkmark \\
\rowcolor{Gray}
print on printer & \texttt{HasPrerequisite} & get printer & \checkmark \\
\rowcolor{Gray}
play game & \texttt{HasPrerequisite} & have game & \checkmark \\
live & \texttt{HasLastSubevent} & die & \checkmark \\
\rowcolor{Gray}
swim & \texttt{HasSubevent} & get wet & \checkmark \\
sit down & \texttt{MotivatedByGoal} & you be tire & \checkmark \\
\rowcolor{Gray}
all paper & \texttt{ReceivesAction} & recycle & \checkmark \\
chair & \texttt{MadeOf} & wood & \checkmark \\
\rowcolor{Gray}
earth & \texttt{DefinedAs} & planet & \checkmark \\
\bottomrule
\end{tabular}
}
\caption{\textbf{Randomly selected and novel} generations from the ConceptNet development set. Novel generations are $sro$ tuples not found in the training set. Manual evaluation of each tuple indicates whether the tuple is considered plausible by a human annotator}
\label{tab:cn_generations}
\end{table}

\section{Related Work}
\label{sec:related}

\paragraph{Knowledge base construction}
Previous work has looked at constructing knowledge bases as relational schemas using expert knowledge \cite{lenat1995cyc,Bodenreider2014,Miller1995},
semi-structured text extraction \cite{Suchanek2007Yago,Hoffart2013,Auer2007DBpediaAN,Bollacker2008FreeBase} and unstructured text extraction \cite{Dong2014,Carlson2010,Nakashole2011,Nakashole2012,Niu2012}. In our work, we focus on construction of commonsense knowledge bases which require the use of open-text events rather than a well-defined relational schema structure.  Other work in information extraction can also be applied to knowledge base construction with open-text entities \cite{Soderland2010AdaptingOI,Etzioni2011OpenIE,fader2011identifying,Mausam2012OpenLL,Fan2010,Cui2018NeuralOI}, but these methods typically extract explicitly stated text relations. Conversely, our approach generates new knowledge that is often unstated in text, as commonsense information typically is \cite{gordon2013reporting}.

\paragraph{Commonsense knowledge base completion} Existing work on generation of novel commonsense knowledge has also used ConceptNet and \textsc{Atomic} as underlying KBs. Specifically, \citet{li2016commonsense} proposed a set of neural network models for scoring tuples in ConceptNet. 
Our work differs from this approach as their models evaluate full tuples rather than learning to generate the phrases to make new nodes in the knowledge graph.
\citet{saito2018commonsense} builds upon this work by proposing a joint model for completion and generation of commonsense tuples. Their work, however, focuses on using tuple generation to augment their KB completion model, rather than to increase coverage in commonsense KB construction. 
Finally, \citet{sap2018atomic} use LSTM encoder-decoder models to generate commonsense knowledge about social situations. We use transformers and investigate the effect of using pre-trained language representations \cite{openaigpt} to initialize them.

\paragraph{Transformers and pre-training} Finally, our work builds on previous work on adapting pre-trained language models for various sequence labeling, classification, and NLI end tasks \cite{openaigpt,elmo,bert}. Our research investigates how pre-trained language models can be used for large-scale commonsense KB construction by generating new graph nodes and edges between nodes. 
\section{Conclusion}

We introduce COMmonsense Transformers (\modelname) for automatic construction of commonsense knowledge bases. \modelname~is a framework for adapting the weights of language models to learn to produce novel and diverse commonsense knowledge tuples. Empirical results on two commonsense knowledge bases, \textsc{Atomic} and ConceptNet, show that \modelname~ frequently produces novel commonsense knowledge that human evaluators deem to be correct. These positive results point to future work in extending the approach to a variety of 
other types of knowledge bases, as well as investigating whether \modelname~can learn to produce OpenIE-style knowledge tuples for arbitrary knowledge seeds.

\section*{Acknowledgments}
We thank Thomas Wolf, Ari Holtzman, Chandra Bhagavatula, Peter Clark, Rob Dalton, Ronan Le Bras, Rowan Zellers and Scott Yih for helpful discussions over the course of this project, as well as the anonymous reviewers for their insightful comments.  This research was supported in part by NSF (IIS-1524371, IIS-1714566, NRI-1525251), DARPA under the CwC program through the ARO (W911NF-15-1-0543), and Samsung Research. This material is based, in part, upon work supported by the National Science Foundation Graduate Research Fellowship Program under Grant No. DGE-1256082.
\bibliography{acl2019}
\bibliographystyle{acl_natbib}

\newpage
\appendix

\section{Additional Training Details}
\label{app:train}

\subsection{Training Hyperparameters}
\label{app:train:hyper}
\paragraph{\textsc{Atomic}}

For \textsc{Atomic}, we use a maximum learning rate of 6.25e-5 with a warmup period of 100 minibatches. After, we decay the learning rate linearly until the end of training. We train for 50k minibatches and use early stopping. We clip gradients when their norm is greater than 1. The remainder of our hyperparameters are the same as in \citet{openaigpt}. We use the public HuggingFace implementation of the GPT model as a base for our experiments available at: \url{https://github.com/huggingface/pytorch-openai-transformer-lm}.

\paragraph{ConceptNet} For ConceptNet, we use a maximum learning rate of 1e-5 and a warm-up period of 200 minibatches.  The learning rate is decayed linearly until the end of training, which lasts for 100k minibatches. All other hyperparameters are the same as for training on the \textsc{Atomic} corpus.

\subsection{ConceptNet baseline} 
We train the ConceptNet baseline with a learning rate of 1e-4 for 100k minibatches. Early stopping is used with the validation loss. Similarly to \citet{saito2018commonsense}, we use 200-dimension hidden states and 200-dimensional word embeddings. We use a single-layer bidirectional LSTM \cite{hochreiter1997long} to encode the first phrase and a single-layer unidirectional LSTM to decode the target phrase. Relation embeddings are concatenated with the word embeddings of the decoder before being input to the decoder LSTM. We set the dropout rate to 0.2 before the output projection layer and after the word embedding layers. We outline the following differences between our re-implementation of the model of \citet{saito2018commonsense} and their original implementation and the reason for the change.

\begin{enumerate}
    \item We use Glove \cite{Pennington2014GloveGV} embeddings rather than fastText embeddings \cite{Bojanowski2017EnrichingWV} to initialize word embeddings. Because the model indicated that 200-dimensional word embeddings were used, we could not use the pretrained embeddings provided by the fastText group\footnote{https://fasttext.cc/}. In  \citet{saito2018commonsense}, the authors described training their fastText embeddings on Wikipedia. With no reference to the precise corpus used, we opted to use Glove embeddings to initialize the word embeddings of the encoder and decoder instead.
    \item We use the Adam optimizer with learning rate of 0.0001, rather than SGD with a learning rate of 1.0 because after training both models, we found that the Adam-trained model performed better on development set perplexity. We also do not use weight decay, as this seemed to lower validation performance, as well.
    \item We do not train the generation model jointly with the completion model. We only train an individual generator. The results of \citet{saito2018commonsense} did not show a significant difference in generation performance between the two on the ConceptNet dataset.
    \item We train a second baseline (LSTM - $s$) that does not learn to produce relations in both directions (i.e., $sr \rightarrow o$ and $or \rightarrow s$). Instead if only learns parameters that can produce relations in the forward direction ($sr \rightarrow o$)
    \item We do not decay the learning rate because it was unclear from the original paper what the exact learning rate schedule was.
\end{enumerate}
\label{app:train:baseline}

\section{Additional Evaluation Details}
\label{app:evaluation}
\subsection{Human Evaluations}
\label{app:human}
We used Amazon Mechanical Turk to get ratings of  model output accuracy.  We selected seed concepts and relations from the test set and generated completions using each model to create $(s, r, o)$ tuples. 
For \textsc{Atomic}, we selected tuples by choosing all possible relations (9) for each of 100 randomly selected seed concepts (900 total $(s,r)$ pairs) following the procedure from \citet{sap2018atomic}. For ConceptNet, we used the full test set (1200 total $(s,r)$ pairs).

For Beam-2/5/10 and top-5/10 sampling generations, we used the model to generate 2, 5, or 10 (respectively) possible  completions ($o$) per $(s,r)$ pair.  Workers were shown the full set and asked to select all of the $o$ that are valid completions for the $(s,r)$ pair. Each set of tuples was rated by 5 workers.

For greedy sampling generations, we used the model to generate one possible completion ($o$) per $(s,r)$ pair.  Workers were shown the completed tuple $(s,r,o)$ and asked whether it is valid or not. Each tuple was rated by 5 workers.

We measure accuracy as the percentage of distinct worker responses where the $(s,r,o)$ tuple is marked as valid (i.e., $\frac{\#valid}{5\cdot|(s,r,o)|}$).

\section{Example Outputs}
\label{app:examples}

\begin{figure*}[t]
    \centering
    \includegraphics[width=\textwidth]{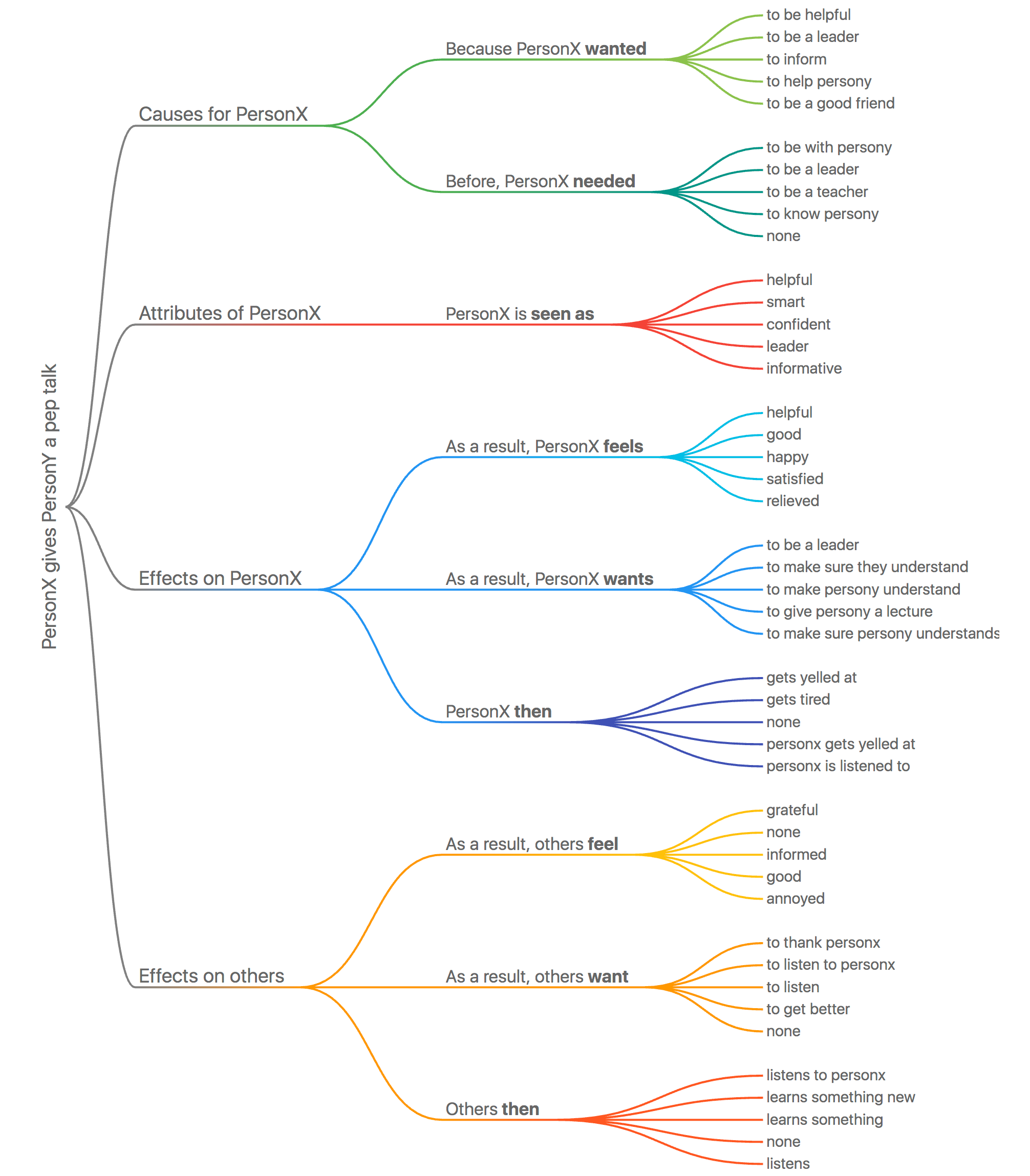}
    \caption{Example outputs for the event "PersonX gives PersonY a pep talk" from \modelname~trained on the \textsc{Atomic} knowledge graph}
    \label{fig:app_example_1}
\end{figure*}

\begin{figure*}[t]
    \centering
    \includegraphics[width=\textwidth]{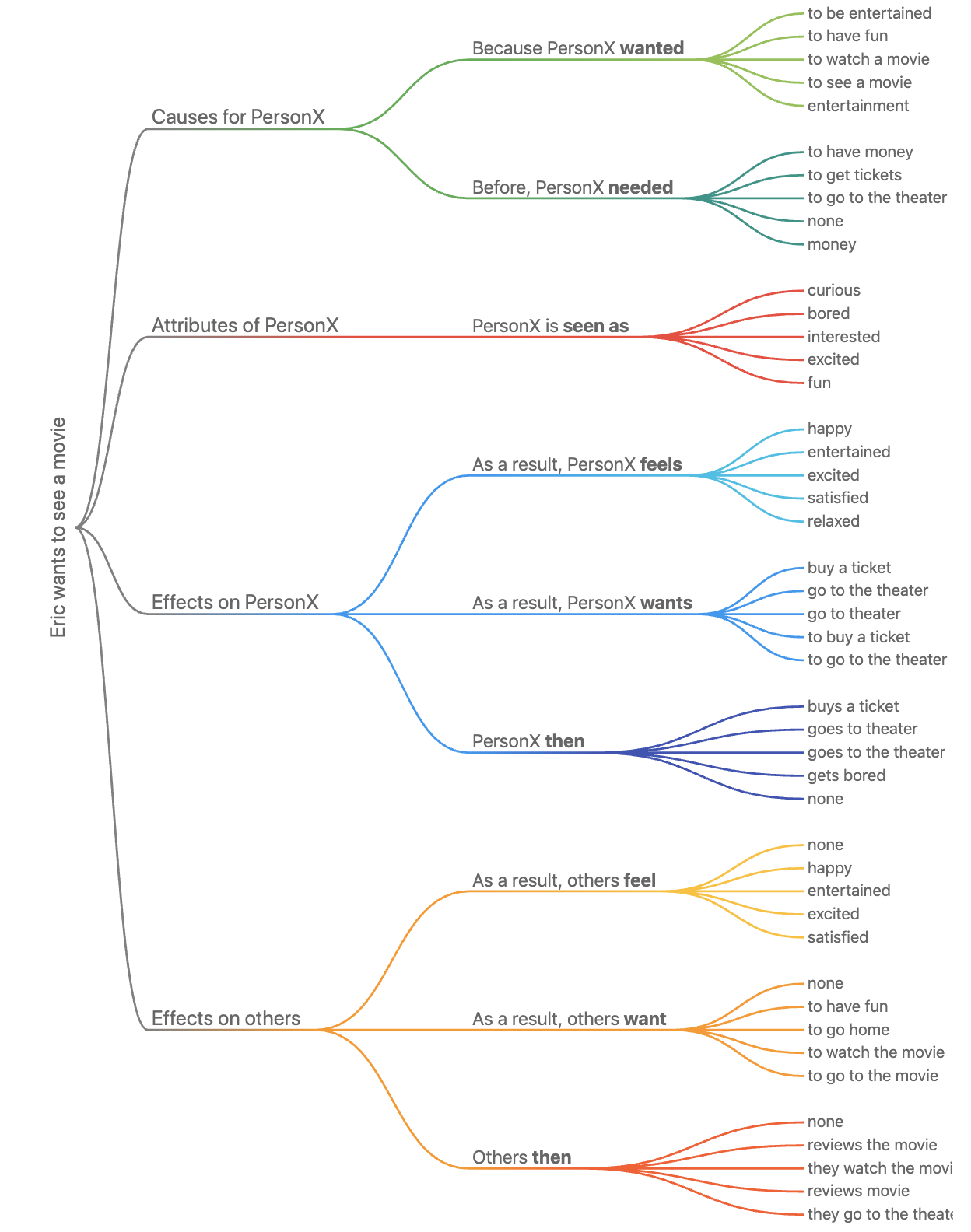}
    \caption{Example outputs for the event "Eric wants to see a movie" from \modelname~trained on the \textsc{Atomic} knowledge graph. \modelname~is able to generalize beyond the templates of the \textsc{Atomic} knowledge graph (i.e., PersonX) and can be used directly with names.}
    \label{fig:app_example_2}
\end{figure*}

\begin{figure*}[t]
    \centering
    \includegraphics[width=\textwidth]{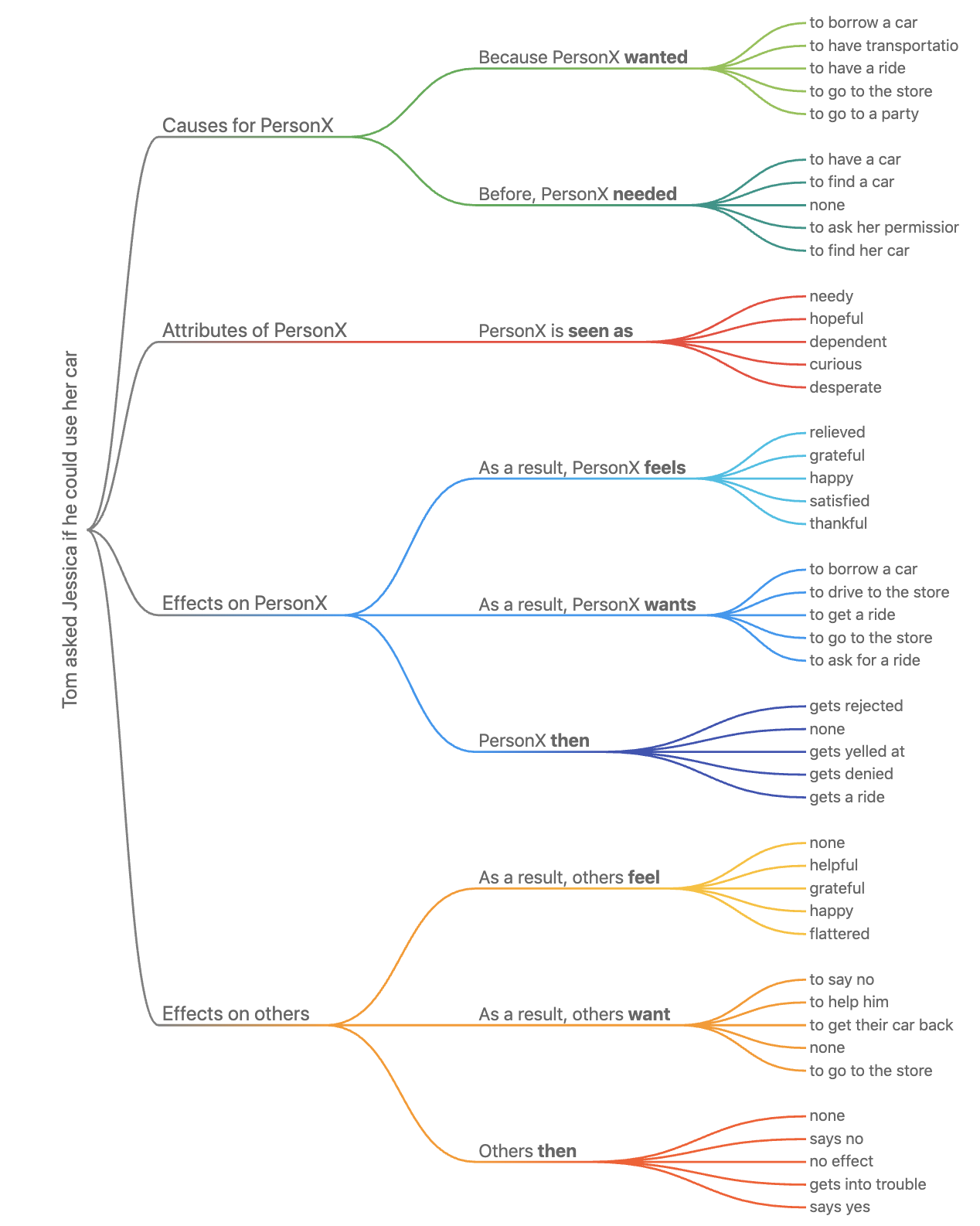}
    \caption{Example outputs for the event "Tom asked Jessica if he could use her car" from \modelname~trained on the \textsc{Atomic} knowledge graph}
    \label{fig:app_example_3}
\end{figure*}

Additional examples can be seen in Figures~\ref{fig:app_example_1}, \ref{fig:app_example_2}, and \ref{fig:app_example_3} that are produced using the demo at \url{https://mosaickg.apps.allenai.org}.

\section{Additional Training Experiments}
\label{app:atomic:train}
In addition to the more naive setups for knowledge graph completion, we explore various multi-task and hierarchical learning setups on top of the taxonomy of commonsense relations given by \citet{sap2018atomic}, which group together along various axes (e.g., related to agent/theme, related to causes/effects, etc.).

\subsection{Multi-relation Training}
\label{app:train:relation}

For the \textsc{Atomic} corpus, we experiment with multiple multi-task training setups, similar to \citet{sap2018atomic}. First, we train an individual model for each relation type (\texttt{oReact, oEffect}, etc.), which we denote as \modelname~- \textsc{9LM} in the Table~\ref{tab:app:multi}. We also experiment with various information-sharing dataset configurations that organize different relations across common dimensions. We outline these dimensions and the makeup of each split in Table~\ref{tab:app:multi}. For ConceptNet, all models are always trained on all relation types jointly. Results on automatic evaluation metrics are provided in Table~\ref{tab:app:auto}. Because there did not seem to be significant differences between these performances and that of \modelname~- \textsc{Full}, we did not run additional experiments on these ablations.

\subsection{Concept Hierarchy Training}
\label{app:train:hierarchy}

Leveraging the prior knowledge that certain relation types in the \textsc{Atomic} knowledge graph are linked to each other, we explore providing these group identities as additional tokens in the relation. For example, when generating the completion of a \texttt{xReact} relation, the model would receive as input the following meta-tokens: \texttt{<xReact>}, \texttt{<X>}, \texttt{<POST>}, \texttt{<Involuntary>} -- thereby providing common context with other relations that are part of the same groupings (e.g., generating a phrase for a \texttt{xWant} relation would receive the \texttt{<X>} and \texttt{<POST>} tokens as input, but not \texttt{<Involuntary>}). Depending on the relation for a particular training example (e.g., \texttt{xReact}), a set of meta-tokens are appended to the relation tokens, $X^r$, that provide hierarchical relational information, allowing the model to share information across relation types. We provide a more in-depth description of the category hierarchy training combinations in Table~\ref{tab:app:hierarchy}. Results on human evaluation metrics are provided in Table~\ref{tab:app:human}. Because the model with the hierarchical meta-tokens performed worse than the regular \modelname, we did not run additional experiments on this ablations.
\newpage
\begin{table*}[t]
\centering
\small
\begin{tabular}{p{2.0cm}p{4.7cm}p{5.5cm}}
\toprule

 \textbf{Event} & \textbf{Description} & \textbf{Example Completion:}     \\
 \midrule
 & & Person X puts Person X's trust in Person Y \\
\midrule 
\texttt{oEffect}    & The effect the event has on others besides Person X &  is considered trustworthy \newline is believed \newline gains Person X's loyalty  \\
\midrule
\texttt{oReact}     & The reaction of others besides Person X to the event  &  trusted\newline honored\newline trustworthy \\
\midrule
\texttt{oWant}      &  What others besides Person X may want to do after the event & work with Person X\newline partner with Person X\newline to help Person X \\
\midrule
\texttt{xAttr}      &  How Person X might be described given their part in the event &  faithful\newline hopeful\newline trusting\\
\midrule
\texttt{xEffect}    &  The effect that the event would have on Person X & gets relieved\newline stays faithful\newline Is betrayed\\
\midrule
\texttt{xIntent}    & The reason why X would cause the event & to be trusting\newline his or her help/guidance/advice\newline to be friends \\
\midrule
\texttt{xNeed}      &  What Person X might need to do before the event & to be friends with Person Y\newline to have heard a lot of good things about Person Y\newline to get to know Person Y\\
\midrule
\texttt{xReact}     &  The reaction that Person X would have to the event &  trusting\newline safe, not alone \newline understood\\
\midrule
\texttt{xWant}      &  What Person X may want to do after the event& to rely on Person Y\newline to go into business with Person Y\newline to make sure that their heart feeling is right \\
\midrule
\end{tabular}
\caption{Definitions of the relations in \textsc{Atomic}. Events in \textsc{Atomic} center around the personal situations of a central figure, Person X, with potentially more participants.}
\label{tab:app:dimensions}
\end{table*}

\begin{table*}[h]
\centering
\small
\begin{tabular}{p{1.6cm}p{5cm}p{7cm}}
\toprule
 Organization & Description & Relations     \\
\midrule 
\textsc{Person X/Y} & The training set is split into relations for the subjects of the event (Person X) and relations for other participants in the event  &   $T_1 =$ \{\texttt{xAttr, xEffect, xIntent},\newline \texttt{xNeed, xReact, xWant}\} \newline $T_2 =$  \{\texttt{oEffect, oReact, oWant}\}  \\
\midrule
\textsc{Pre/Post} & Event preconditions are jointly trained (i.e., intentions, needs). Event postconditions are jointly trained.  &   $T_1 =$  \{\texttt{xIntent, xNeed}\} \newline $T_2 =$ \{\texttt{oEffect, oReact, oWant},\newline \texttt{xEffect, xReact, xWant}\}   \\
\midrule
\textsc{(In)Volun} & Involuntary relations are trained jointly, such as reactions and effects. Voluntary relations are trained jointly, such as needs, wants, and intents. &  $T_1 = $ \{\texttt{oWant, xIntent, xNeed, xWant}\} \newline $T_2 = $ \{\texttt{oEffect, oReact, xAttr}, \newline \texttt{xEffect, xReact}\}  \\
\midrule
\textsc{Full} & The training set is made up of all relations and the model is trained jointly on all of them & $T_1 =$ \{\texttt{oEffect, oReact, oWant, xAttr}, \newline \texttt{xEffect, xIntent, xNeed, xReact, xWant}\} \\
\bottomrule
\end{tabular}
\caption{Multi-relation training setups. Following \citet{sap2018atomic}, the \texttt{xAttr} relation is not included in the \textsc{Pre/Post} training configuration}
\label{tab:app:multi}
\end{table*}

\begin{table*}[h]
\centering
\small
\begin{tabular}{p{2.6cm}p{5cm}p{6cm}}
\toprule
 \textbf{Meta-Token} & \textbf{Description} & \textbf{Relations}     \\
\midrule 
\texttt{<X>} & Appended to relations that describe an attribute of Person X  &  \texttt{xAttr, xEffect, xIntent, xNeed, xReact, xWant} \\
\midrule
\texttt{<Y>} & Appended to relations that describes an attribute of a participant that is not Person X  &  \texttt{oEffect, oReact, oWant} \\
\midrule
\texttt{<Pre>} & Appended to relations that correspond to pre-conditions of the event  &   \texttt{xIntent, xNeed}  \\
\midrule
\texttt{<Post>} & Appended to relations that correspond to post-conditions of the event  &  \texttt{oEffect, oReact, oWant, xEffect, xReact, xWant}   \\
\midrule
\texttt{<Voluntary>} & Appended to relations that correspond to voluntary dimensions of the situation &  \texttt{oWant, xIntent, xNeed, xWant} \\
\midrule
\texttt{<Involuntary>} & Appended to relations that correspond to involuntary dimensions of the situation &  \texttt{oEffect, oReact, xAttr, xEffect, xReact}  \\
\bottomrule
\end{tabular}
\caption{Category hierarchy meta-tokens, along with the description and the relations to which they are appended}
\label{tab:app:hierarchy}
\end{table*}

\begin{table*}[t]
\centering
\resizebox{0.8\linewidth}{!}{
\begin{tabular}{l rrr  rrr  rrr r}
\toprule
 \textbf{Model} & \textbf{PPL}\footnotemark[3] & \textbf{BLEU-2} &  \textbf{N/T $sro$}\footnotemark[4] & \textbf{N/T $o$} & \textbf{N/U $o$}  \\
 \toprule
 \modelname - \textsc{9LM} &11.72 & 14.89 &  100.00 & 9.45 & 49.89  \\
  \modelname - \textsc{(In)Volun} & 11.38 & 14.99 &  100.00          & 8.60& 48.36   \\
 \modelname - \textsc{PersonX/Y} & 11.30 & 15.21 & 100.00            & 9.12& 49.59  \\
 \modelname - \textsc{Pre/Post}  & 11.35 & 14.88 & 100.00            & 9.86& 51.86   \\
 \midrule
 \modelname - \textsc{Full} (- pretrain) & 15.42 & 13.88 & 100.00      & 7.25& 45.71  \\
 \modelname - \textsc{Full} & 11.14 & 15.10 &  100.00                      & 9.71 & 51.20   \\
\modelname - \textsc{Full} (+ hierarchy meta-tokens) & \textbf{10.98} & \textbf{15.27} & 100.00  & \textbf{10.03} & \textbf{51.97}\\

\bottomrule
\end{tabular}}
\caption{Automatic evaluations of quality and novelty for generations of \textsc{Atomic} commonsense that are trained with the training set split along different relation types. The training splits are outlined in Table~\ref{tab:app:multi}.}
\label{tab:app:auto}
\end{table*}

\definecolor{lightgray}{rgb}{0.95, 0.95, 0.95}
\newcolumntype{g}{>{\columncolor{lightgray}}c}

\begin{table*}[t]
\centering
\resizebox{\linewidth}{!}{
\begin{tabular}{l || rrr  rrr  rrr ||g}
\toprule
 \textbf{Model} & \textbf{oEffect} & \textbf{oReact} & \textbf{oWant} & \textbf{xAttr} & \textbf{xEffect} & \textbf{xIntent} & \textbf{xNeed} & \textbf{xReact} & \textbf{xWant} & \textbf{Total} \\ 
 \toprule
\modelname & \textbf{29.02} & 37.68 & \textbf{44.48} & \textbf{57.48} & \textbf{55.50} & \textbf{68.32} & \textbf{64.24} & \textbf{76.18} & 75.16 & \textbf{56.45} \\
\modelname~(+ hierarchy meta-tokens) & 28.46 & \textbf{38.96} & 43.64 & 51.90 & 50.84 & 63.00 & 63.98 & 66.20 & \textbf{75.82} & 53.64 \\
\bottomrule
\end{tabular}}
\caption{Human score of generations of \textsc{Atomic} commonsense for the regular \modelname~model and the \modelname~+ category meta tokens}
\label{tab:app:human}
\end{table*}

\end{document}